\documentclass[conference]{IEEEtran}
\usepackage{cite}
\usepackage{amsmath,amssymb,amsfonts}
\usepackage{graphicx}
\usepackage{textcomp}
\usepackage{url}
\usepackage{algorithm}
\usepackage{algpseudocode}
\usepackage{xcolor,subfigure,multirow}
\usepackage{orcidlink,soul}

\def\BibTeX{{\rm B\kern-.05em{\sc i\kern-.025em b}\kern-.08em
    T\kern-.1667em\lower.7ex\hbox{E}\kern-.125emX}}

\graphicspath{{images/}}
\begin{document}
\title{Using Machine Learning to Detect Rotational Symmetries from Reflectional Symmetries\\ in 2D Images}

\author{\IEEEauthorblockN{Koen Ponse \orcidlink{0000-0002-6542-3042}}
\IEEEauthorblockA{\textit{LIACS, Leiden University}\\
Leiden, The Netherlands \\
kpponse@gmail.com }
\and
\IEEEauthorblockN{Anna V. Kononova \orcidlink{0000-0002-4138-7024}}
\IEEEauthorblockA{\textit{LIACS, Leiden University}\\
Leiden, The Netherlands \\
a.kononova@liacs.leidenuniv.nl}
\and
\IEEEauthorblockN{Maria Loleyt}
\IEEEauthorblockA{\textit{Ornamika}\\
Moscow, Russia\\
maria\_loleyt@ornamika.com}
\and
\IEEEauthorblockN{Bas van Stein \orcidlink{0000-0002-0013-7969}}
\IEEEauthorblockA{\textit{LIACS, Leiden University}\\
Leiden, The Netherlands\\
b.van.stein@liacs.leidenuiv.nl}
}

\maketitle
\begin{abstract}
Automated symmetry detection is still a difficult task in 2021.
However, it has applications in computer vision, and it also plays an important part in understanding art. 
This paper focuses on aiding the latter by comparing different state-of-the-art automated symmetry detection algorithms.
For one of such algorithms aimed at reflectional symmetries, we propose post-processing improvements to find localised symmetries in images, improve the selection of detected symmetries and identify another symmetry type (rotational).
In order to detect rotational symmetries, we contribute a machine learning model which detects rotational symmetries based on provided reflection symmetry axis pairs. We demonstrate and analyze the performance of the extended algorithm to detect localised symmetries and the machine learning model to classify rotational symmetries.
\end{abstract}

\begin{IEEEkeywords}
symmetry detection, machine learning, rotational symmetry, reflection symmetry
\end{IEEEkeywords}

\section{Introduction}
In everyday life, humans and animals use their innate perception of symmetry to identify objects quickly. Also in computer vision, symmetry plays an important role in understanding objects. For example, the knowledge that an object is symmetrical, such as a car, means that we do not need to observe the entire object in order to know how it looks like.
Symmetry also plays a significant role in art as it is one of the main principles with which artists create the composition of an artwork \cite{phdthesis}.
Automated symmetry detection can therefore improve various computer tasks and aid art scientists and cultural heritage projects. 
In this paper we aim to help the latter by analysing the current state-of-the-art automated symmetry detection algorithms on applied art data. Furthermore, an algorithm extension is proposed to find local reflection symmetries and a machine learning model is proposed to detect rotational symmetries using reflection symmetry pairs.

Online archive of world ornaments and patterns \textit{Ornamika} \cite{ornamika} has kindly provided us with their database containing a set of diverse images of art pieces.
Labelling these images with symmetries and symmetry types will help organise their database. This will, in turn, aid designers using the database. Due to a large number of images in the database, an automated approach for labelling the images is required.
The Ornamika data set differs from most data sets on which existing symmetry detection algorithms have been tested -- most papers test their algorithms on data sets containing real-world photos with complex backgrounds, lighting and depth.
In contrast, the Ornamika data set is mainly comprised of ornaments on a 2D plane. Often the image features a flat textile art piece without any irrelevant background objects. When an image in the Ornamika data set is not a flat surface, it is often an object which has been photographed \textit{en face} in good lighting conditions. Since other papers have used a different kind of data, it is currently unknown which algorithms perform best on a data set like Ornamika's. 
Additionally, it is difficult to predict whether existing algorithms can be improved to yield better results. 

Over the years, many papers \cite{atallah1984symmetry,loy2006detecting,cho2009bilateral,ming2013symmetry,sympascal-ke2017srn} have been published proposing various symmetry detection methods.
Up until now, three symmetry detection competitions have been held, with increased task complexity in each competition. Further details about these competitions are provided in Section \ref{relatedwork}.
Despite the amount of work in the field, automated symmetry detection does still prove to be a difficult task, as we can conclude based on the last symmetry detection competition in 2017 \cite{comp3}. At the same time, an algorithm proposed in 2006 is still considered competitive.

In this paper we discuss the current state-of-the-art algorithms for symmetry detection in Section \ref{relatedwork}. Next, we compare these algorithms on the Ornamika data set to determine which algorithm works best on such art data. A global reflection symmetry detection algorithm \cite{elawady2017wavelet} is selected from this comparison and several improvements are proposed in Section \ref{methodology}.

We propose to use the algorithm output in order to detect rotational symmetries using a machine learning model (Section \ref{ml_approach}) and we show that our model is able to detect rotational symmetries with an accuracy of over 90\% on the test set.

We note that, in theory, any reflection symmetry detection algorithm will be able to use our trained model to detect rotational symmetries, as its input only requires a set of coordinates of two lines and their respective symmetry scores. Most reflection symmetry detection algorithms produce these attributes to draw symmetry lines.
Our final algorithm is validated on 200 images. The results of this, together with the source code, can be found on GitHub\footnote{\url{https://github.com/Koen-Git/ColorSymDetect}}.

\section{Types of Symmetry}

\begin{figure*}[ht]
    \centering
    \subfigure[Reflection]{\includegraphics[width=0.16\linewidth,trim={0mm 0mm 0mm 0mm},clip]{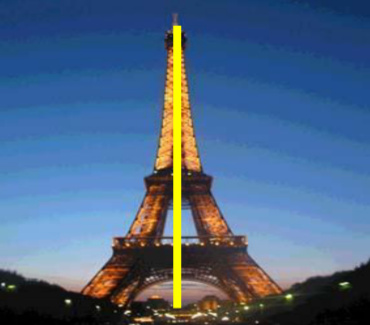}\label{fig:reflection}}
    \subfigure[Rotation]{\includegraphics[width=0.16\linewidth,trim={0mm 0mm 0mm 0mm},clip]{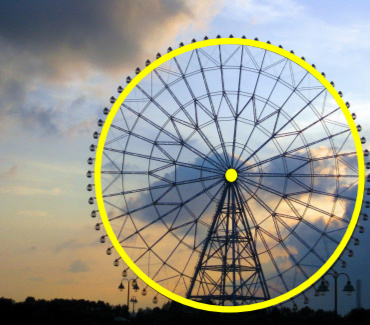}\label{fig:rotation}}
    \subfigure[Translation]{\includegraphics[width=0.16\linewidth,trim={0mm 0mm 0mm 0mm},clip]{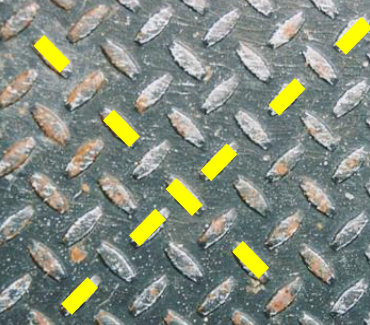}\label{fig:translation}}
    \subfigure[Glide-reflection]{\includegraphics[width=0.16\linewidth,trim={0mm 0mm 0mm 0mm},clip]{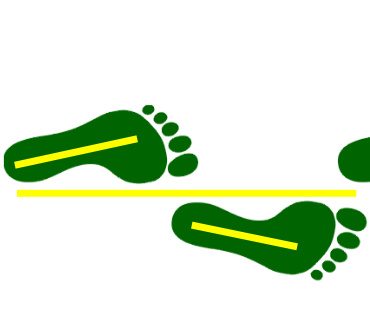}\label{fig:glide}}
    \caption{The four types of symmetry with the symmetry axis shown in yellow (source: \cite{phdthesis}).}
    \label{fig:symtypes}
\end{figure*}

Symmetries can be categorised into four different types, namely \cite{phdthesis}:
\begin{itemize}
    \item Reflection
    - where the image appears mirrored alongside a straight axis. Figure \ref{fig:reflection} shows an example where the left half of the image is a mirrored version of the right half.
    \item Rotational 
    - where elements are repeated around a centre point. Figure \ref{fig:rotation} shows an example where all the spikes of the wheel are centred around the middle of the ferris wheel.
    \item Translation
    - where elements are repeated while rotated in a different direction. Figure \ref{fig:translation} displays an example where the metallic street stones are repeated in a different direction.
    \item Glide reflection
    - is a combination of reflection symmetry and translation symmetry. Figure \ref{fig:glide} shows an image where the footprints are mirrored along a straight axis and are also translated in a forward direction.
\end{itemize}

\section{Related Work}\label{relatedwork}
Three different symmetry detection competitions were held in 2011, 2013 and 2017 \cite{comp1, comp2, comp3}. These symmetry detection competitions featured state-of-the-art algorithms of their respective times. The first two symmetry competitions featured reflection, rotation and translation symmetries. In both the reflection and rotation categories Loy and Eklundh \cite{loy2006detecting} came out as the winner.
The latest symmetry detection competition, held in 2017, featured the medial axis detection category for the first time. Loy and Eklundh \cite{loy2006detecting} still acted as the baseline for reflection and rotational symmetries. A second baseline algorithm \cite{sie2013detecting} for reflection symmetry was introduced and outperformed Loy and Ekhlundh on single reflection symmetries. Surprisingly Loy and Eklundh once again won in detecting multiple symmetries, with Elawady et al. \cite{elawady2017wavelet} coming in second place. For translation symmetries, the scores of the challenger \cite{michaelsen2017hierarchical} and baseline \cite{wu2010detecting} algorithms were still relatively low, with $F_1$-scores\footnote{$F_1$-score defined in (\ref{eq1}) offers a single-value score for system performance comparison, by using the precision and recall \cite{comp3}} 
of 0.19 and 0.20 respectively. In the new category (medial axis) two algorithms managed to beat the baseline algorithm. Liu et al. \cite{liu2017fusing} came in first place with an $F_1$-score of 0.73. For comparison, humans achieved an $F_1$-score of 0.80 \cite{comp3}. 
Funk and Liu \cite{funk2017beyond} acted as a second baseline in the rotational symmetry category and the authors themselves claim better performance over Loy and Eklundh. However, Funk and Liu only focus on finding the rotational centres. Furthermore, Funk and Liu produces a dense heatmap of possible symmetries instead of coordinates of possible symmetries.

Since then, Elawady has evaluated different versions of his algorithm on seven different symmetry data sets, four for single symmetry detection and three for multiple axis detection \cite{phdthesis} -- the results are listed in Table \ref{tab:ela_own_table}. These results show that Elawady's algorithm has a slight edge in the last competition's multiple axis training data set (ICCV17m). The test set of this same competition shows Loy has the higher performance \cite{comp3}. Nonetheless, when considering the other 6 data sets, we can see Elewady has a clear advantage in each of them over Loy and Eklundh.

\begin{table}[]
    \centering
    \caption{Elawady algorithm compared to Loy and Eklundh and Cicconet et al. on seven different data sets. Each cell lists the amount of top-ranked detected symmetry axis that corresponds to a ground truth in the data set. Between brackets the $\max\{F1\}$ scores are listed (data source: \cite{phdthesis}).}
    \begin{tabular}{|c|c|c|c|}
    \hline
        Data set (\#images) & Loy \cite{loy2006detecting} & Cicconet \cite{cicconet2014mirror} & Elawady \cite{elawady2017wavelet} \\
        \hline
        \hline
        PSU (157) & 81 (0.514) & 90 (0.569) & \textbf{113} \textbf{(0.724)} \\
        AVA (253) & 174 (0.690) & 124 (0.493) & \textbf{182} \textbf{(0.729)} \\
        NY (176) & 98 (0.528) & 92 (0.526) & \textbf{135} \textbf{(0.766)} \\
        ICCV17 (100) & 52 (0.507) & 53 (0.536) & \textbf{70} \textbf{(0.713)} \\
        \hline
        PSUm (142) & 69 (0.292) & 68 (0.159) & \textbf{75} \textbf{(0.338)} \\
        NYm (63) & 32 (0.337) & 36 (0.237) & \textbf{40} \textbf{(0.411)} \\
        ICCV17m (100) & 54 (0.273) & 39 (0.207) & \textbf{57} \textbf{(0.285)} \\
        \hline
        \hline
        Total (991) & 560 (0.449) & 502 (0.390) & \textbf{672} \textbf{(0.567)} \\
        \hline
    \end{tabular}
    
    \label{tab:ela_own_table}
\end{table}

We can note that $F_1$-scores are still very low during the last competition for multiple and single reflection symmetry detection: $F_1$-scores of 0.3 and 0.52 respectively. Elawady's results show higher $F_1$-scores: 0.73 for single reflection data sets and 0.34 for multiple reflection data sets. However, it is clear that there is still much room for improvement in this field.

Detecting rotational symmetries has not received much attention in the last symmetry competition. Loy and Eklundh \cite{loy2006detecting} won this category in the first and second symmetry competition and is considered the best baseline for this category. Funk and Liu \cite{funk2017beyond} claim better performance but only focus on finding the rotational centres in their work. 

\section{Evaluation methods}
In 2019 Elawady published his thesis \cite{phdthesis} in which he states the evaluation method for reflection symmetry used in the competitions is unclear. 
Elawady explains how to evaluate symmetry axis to their corresponding ground truths. 
Just like the competitions, he uses the $\max F_1$ score defined in (\ref{eq1}) as a global quality measure for comparing algorithms. 
The $F_1$ score is the harmonic mean of precision $P$ and recall $R$: 
\begin{equation} \label{eq1}
    F_1 = \frac{2P \cdot R}{P + R}
\end{equation}
Precision $P$ is defined as the number of true positives over the number of true positives plus the number of false positives. While recall $R$ is defined as the number of true positives over the number of true positives plus the number of false negatives.
For evaluation, one can alter thresholds of when a detected symmetry is classified as a true positive. This creates a plot of $F_1$ scores which illustrates the trade off between precision and recall. The maximum in this plot is a convenient performance measurement for detection algorithms. 

\begin{figure}[h!]
    \centering
    \includegraphics[width=0.3\linewidth]{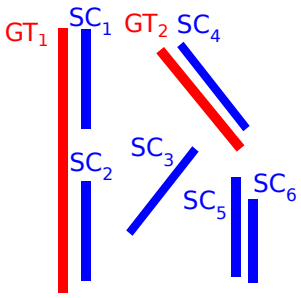}
    \caption{Example with two ground truths (red) and six detected symmetry lines (blue). The detected symmetry lines are ordered by their respective symmetry score produced by the algorithm. SC1 and SC2 are matched to GT1 (the difference in their slope and their centres are within some threshold) and SC4 is matched to GT2. Furthermore, SC6 can be disregarded as it is similar to the higher-ranked SC5. (source: \cite{phdthesis})}
    \label{fig:toy_example}
\end{figure}

Furthermore, Elawady emphasizes that a pair of similar symmetry lines should not both be evaluated. Instead, only the higher ranked symmetry line (with a higher score) should be considered. 
Next, he states that multiple detected symmetry axes can be grouped to form a single ground truth.
Both practices are visualised in Figure \ref{fig:toy_example}.
It is unclear whether the symmetry competitions also adopted these methods, as they only list the precision and recall with their corresponding $F_1$-scores on a single test set.

\section{Methodology} \label{methodology}
Based on our literature review, we conclude that for reflection symmetries, Elawady's WaveletSym algorithm \cite{elawady2017wavelet} achieves the highest performance on the largest distribution of data sets while using a clear evaluation method. Furthermore, we found that WaveletSym processes images in under ten seconds on average. In Section \ref{ml_approach}, we compare the computation time of our work with Loy and Eklundh. We find that Loy and Eklundh's method requires four minutes on average to process an image. This time is comparable to our proposed work in which we recursively use WaveletSym multiple times. For these reasons, we consider WaveletSym currently state of the art for detecting reflection symmetries\footnote{We are aware that the latest version of Elawady's algorithm \cite{phdthesis} is even more sophisticated but since no open-source version is available, we used a slightly older version of the algorithm for our research. Note that this does not impact the performance of the proposed machine learning model}.

In this Section we will examine the performance of WaveletSym on our data set and propose post-processing improvements to this algorithm.
Furthermore, we propose a machine learning model to detect rotational symmetries using the output of WaveletSym and compare the final performance of our model against Loy and Eklundh \cite{loy2006detecting}.

\subsection{Wavelet-based Reflection Symmetry Detection}\label{wavelet-based}
\begin{figure}[h!]
    \centering
    \includegraphics[width=0.85\linewidth]{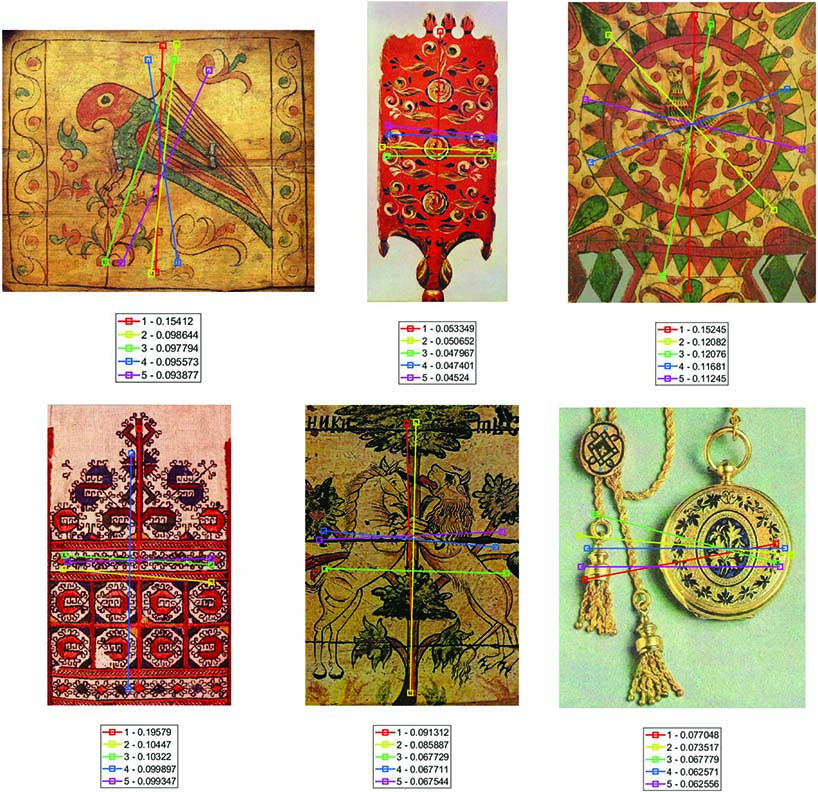}
    \caption{Results as produced by \cite{elawady2017wavelet} \textit{without any alterations}. The top 5 discovered symmetries are drawn on the image. The numbers in the legend represent each lines symmetry score, how strong of a symmetry is present.}
    \label{fig:ElawadyOriginal}
\end{figure}

Elawady's \cite{elawady2017wavelet} WaveletSym algorithm produces a multidimensional array of symmetry lines with their corresponding symmetry score. 
As shown in Figure \ref{fig:ElawadyOriginal}, these lines are then drawn on the image with the legend of the scores attached.

\paragraph{Initial results}
WaveletSym produces pretty satisfying results, as is shown in Figure \ref{fig:ElawadyOriginal}. With default settings, the top 5 symmetries are drawn on the image. We can see WaveletSym often finds the main symmetry axis, with secondary symmetries being found relatively accurate as well. 
We should note that in a data set mainly containing art pieces like the Ornamika \cite{ornamika} data set, the main symmetry axis is often the least interesting. This is because in a vast majority of the images a reflection symmetry exists in the centre of the image.

\paragraph{Further results}
When inspecting the resulting images in Figure \ref{fig:ElawadyOriginal} we observe a few shortcomings of the WaveletSym algorithm in its current state.
\begin{itemize}
    \item As is described in \cite{elawady2017wavelet}, the algorithm only deals with global reflection symmetries, while we are also interested in more localised symmetries. Especially, when considering that the main global symmetry axis is often the least interesting in our data set.
    \item This algorithm only deals with reflection symmetries, but we are also interested in rotational symmetries since many art images contain such symmetries.
    \item All symmetry lines with a score over 0 are returned by the algorithm, with no way to filter on accurate symmetry lines. Yet, the scores themselves are not a very meaningful metric when comparing scores from one image to another.
    \item The algorithm produces many symmetry lines that are very similar to each other.
\end{itemize}

\paragraph{Improvements}
Despite these shortcomings, WaveletSym is still a promising algorithm, and some alterations will mitigate these described problems.
In the next paragraph, we will go through these problems one by one and suggest improvements to mitigate them as much as possible.

\begin{figure}[ht]
    \centering
    \subfigure{\includegraphics[width=0.45\linewidth,trim={20mm 15mm 15mm 15mm},clip]{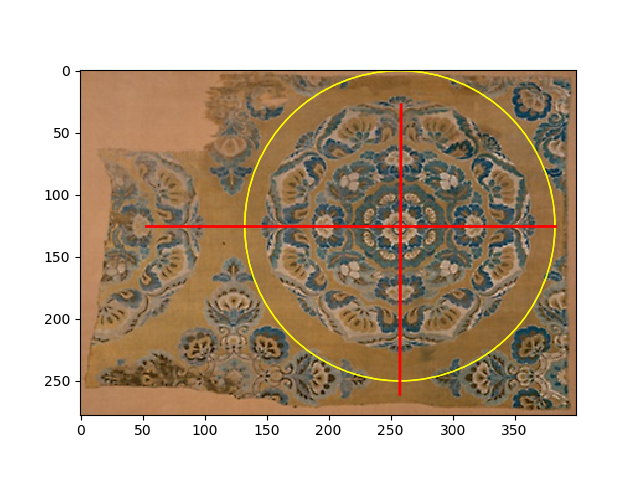}}\label{fig:rot_good}
    \subfigure{\includegraphics[width=0.2\linewidth,trim={54mm 13mm 48mm 13mm},clip]{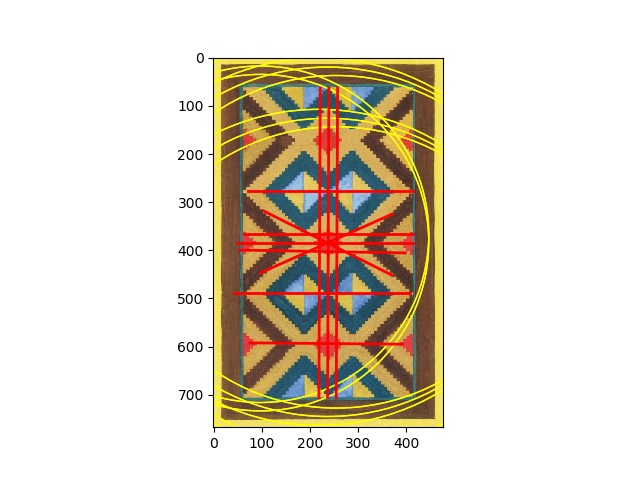}}\label{fig:rot_bad}
    \caption{Rotational symmetries drawn with with a \textit{naive approach}. The red lines represent the discovered reflection symmetries. Any perpendicular lines will draw a yellow rotational symmetry in their intersection.}
    \label{fig:rotational}
\end{figure}

\begin{itemize}
    \item \textbf{Localised symmetries:} We will cut the images at certain points to find localised symmetries in smaller parts of the images. 
    For determining where to cut the image, we could cut at as many places as possible. However, this results in undesired symmetry lines and most notably increases computation time exponentially.
    Instead, we settled on recursively cutting up the image using the top three symmetry lines. This creates six new (smaller) images which can be processed in the same manner. This resulted in a good balance between computation time and retrieving accurate local symmetries. 
    
    \item \textbf{Rotational symmetry:} First, we observed that when two lines cross and are perpendicular, there is likely a rotational symmetry around the intersection as shown in the left subfigure of Figure \ref{fig:rotational}. The rotational symmetries as drawn with this \textit{naive} approach lead to unwanted results, as shown in the right subfigure of Figure \ref{fig:rotational}. These results are partly caused by our earlier described problem of similar symmetry lines being drawn close to each other; this also causes multiple rotational symmetries to be drawn. The other problem is that there are not necessarily rotational symmetries at the intersections between two perpendicular lines.
    Requiring the two perpendicular lines to have similar symmetry scores improves the performance. 
    Besides these simple rules, we contribute a machine learning approach for detecting rotational symmetries with symmetry line pairs as an input. We will elaborate more on this in Section \ref{ml_approach}.
    
    \item \textbf{Filtering:} To determine if symmetries should be drawn on the image, we can use the normalised score for each image or sub-image. First, we check if the top symmetry found reaches our threshold parameter. Then we check if the other detected symmetry scores lie within a second threshold compared to the first one. Symmetries that do not reach these thresholds will be removed.
    
    \item \textbf{Similar symmetries:} Since WaveletSym is producing a lot of similar symmetry lines, we can loop over each line and compare them to each other. If we find lines with a similar slope and centre, we can discard the line with a lower symmetry score.
    The same procedure can be performed on rotational symmetries, i.e. circles drawn with a similar radius and centre.
\end{itemize}

\begin{figure}[ht]
    \centering
    \subfigure{\includegraphics[width=0.82\linewidth,trim={0mm 0mm 0mm 0mm},clip]{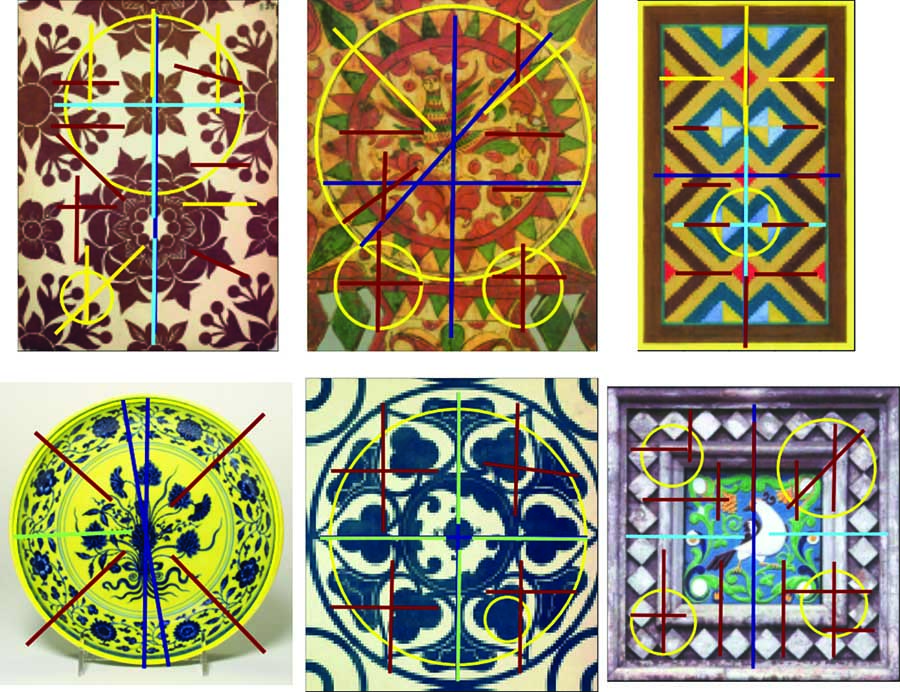}}
    \subfigure{\includegraphics[width=0.038\linewidth,trim={0mm 0mm 0mm 0mm},clip]{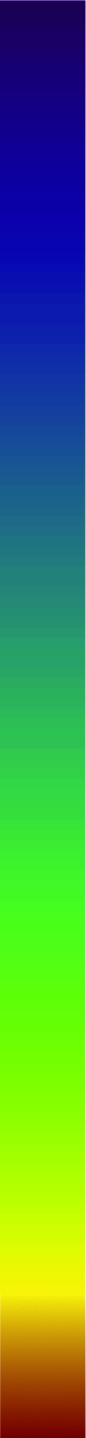}}
    \caption{Resulting images of WaveletSym \textit{with our alterations} outlined in Section \ref{wavelet-based}. The blue lines are the discovered global symmetries. Other colours represent different depths of the recursive function and, in turn, represent discovered localised symmetries. The colours of the different depths are mapped in the colour map to the right. Colours presented lower on the map represent more localised symmetries.} \label{fig:ela_improved}
\end{figure}

\paragraph{Results after improvements}
Figure \ref{fig:ela_improved} shows images with all our proposed adjustments. The rotational symmetries are detected in these images using the described rule-based approach, not using machine learning.
For this set, we have set our first parameter, 'symThresholdAC', to 0.20. This parameter dictates whether the top symmetry of each image or sub-image should be kept or not. Any lines with scores below this parameter are removed. Lines generated as part of the same recursive loop are removed as well.
We have set our second parameter, 'normThresholdAC', to 0.70. This parameter dictates which other symmetry lines are kept. 
Lines with scores relative to the top symmetry score which are lower than this threshold, are removed.
The third parameter, 'CircleSymThreshold', was set to 0.75; this implies the maximum difference in symmetry scores between two lines can at most be 25\% for them to create a rotational symmetry.

The resulting images show that with these adjustments, WaveletSym is capable of finding localised symmetries and rotational symmetries. However, some rotational symmetries are still not found, and some smaller ones are found inaccurately. Additionally, some localised symmetries which are discovered are inaccurate. The performance of this rotational symmetry detection method will function as a good baseline for our proposed machine learning model, which we will introduce next.

\subsection{Machine Learning for Rotational Symmetries}\label{ml_approach}
Discovering rotational symmetries as described in the previous Section \ref{wavelet-based} amounts to reasonable results. In order to improve these results further we propose to use a machine learning model to learn when reflection symmetry lines indicate a rotational symmetry at their intersection.
For this classification task, we first had to label symmetry line combinations manually. We labelled lines that intersected and had a rotational symmetry at the intersection as true. Lines were also required not to be parallel or close to parallel. Other line pairs are labelled as false. In order to balance the data set, we also rotated every true case by 0.5 degrees 720 times.

Thus, our input is comprised of $x$ and $y$ coordinates of two line pairs with their corresponding symmetry score. We pre-process this data into several (hand-crafted) features as the input for our model, namely: 
the difference of the symmetry scores, the slopes of both lines and their difference, the perpendicular slope of one of the lines and the difference of this perpendicular slope to the other line, difference in length of the lines, whether or not the lines intersect, and the difference of each line's endpoint to the intersecting point.

We compared eight different machine learning algorithms trained on the training data and validated on a different set of validation data. These sets contain 153785 and 517213 samples respectively. The results are available in Table \ref{tab:scores_init}\footnote[3]{In Table \ref{tab:scores_init} and \ref{tab:scores_best}: all non listed parameters have default values, using sklearn 0.24.2 (python 3.7.10). Each algorithm, except for K-Nearest Neighbors, was also set with "random\_state=44" and "class\_weight='balanced'".}.

\begin{table}[h]
    \caption{Accuracy scores of various algorithms on validation data. While SVC\_auto achieves the highest accuracy, the ROC curve (Figure \ref{fig:ROC_all}) shows that Random Forest and LinearSVC perform the best in terms of AUC.}
    \centering
    \begin{tabular}{|p{0.20\linewidth}|p{0.35\linewidth}|p{0.15\linewidth}|p{0.10\linewidth}|}
        \hline
         \textbf{Algorithm} & \textbf{Parameters\footnotemark[3]} & \textbf{Accuracy} & \textbf{AUC} \\
         \hline
         \hline
         Log\_Regression & max\_iter=10000 & 0.82181 & 0.50 \\
         Random Forest & & 0.85381 & \textbf{0.94} \\
         SVC &  & 0.19588 & 0.51 \\
         SVC\_alt & gamma="auto" & \textbf{0.88888} & 0.76 \\
         KNN & & 0.86131 & 0.71 \\
         Decision Tree &  & 0.87025 & 0.67 \\
         SGDClassifier &  & 0.46510 & 0.58 \\
         LinearSVC & max\_iter=5000 & 0.88024 & \textbf{0.94} \\
         \hline
    \end{tabular}
    \label{tab:scores_init}
\end{table}

\begin{figure}[htpb]
    \centering
    \includegraphics[width=0.7\linewidth]{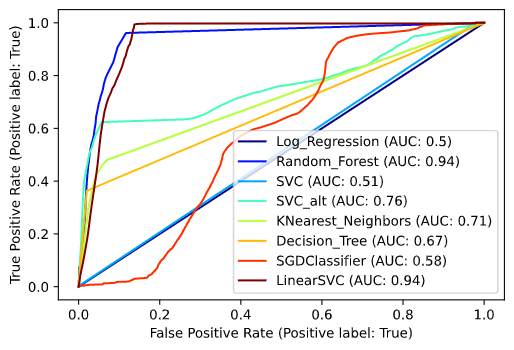}
    \caption{ROC Curve of the scores in Table \ref{tab:scores_init}. Random Forest and LinearSVC come out on top in terms of AUC.}
    \label{fig:ROC_all}
\end{figure}

From these model configurations, the Support Vector Classification (SVC), with its 'gamma' hyperparameter set to "auto" achieved the highest accuracy.
Alternatively to evaluating the accuracy we can instead calculate the true-positive rate $(\frac{True Positives}{True Positives + False Negatives})$ and the false-positive rate $(\frac{False Positives}{False Positives + True Negatives})$. Plotting both of these rates against each other at different threshold values will produce a receiver operating characteristic curve (ROC curve). When plotting the true positive rate on the y-axis of the curve, it is often desired to have a large area under the curve (AUC), as this corresponds with a high true positive rate and a low false positive rate. 
By further evaluating the Support Vector Classification, we find that this algorithm is scoring poorly when considering the area under curve score. Random Forest and LinearSVC achieved better performance with scores of 0.94.

\begin{figure}[htpb]
    \centering
    \includegraphics[width=0.7\linewidth]{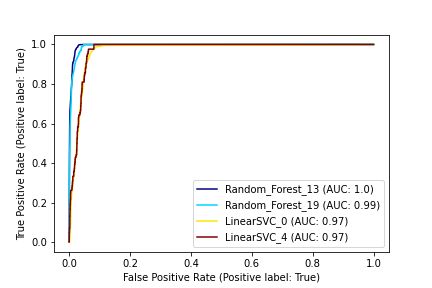}
    \caption{ROC curves of the different hyper parameter configurations listed in Table \ref{tab:scores_best}.}
    \label{fig:ROC_best}
\end{figure}

We re-evaluated the three top-performing algorithms using $36$ different hyperparameter configurations, the results of this experiment can be found on our Github page.
From the hyperparameter optimization step, we evaluated the top-performing Random Forest and LinearSVC configurations one more time on a completely different test set, containing 263569 samples.
The final results of this test are found in Table \ref{tab:scores_best}\footnotemark[3], and the Receiver Operating Characteristic curve is found in Figure \ref{fig:ROC_best}. 
We conclude that Random Forest with "max\_depth=10" and "criterion='entropy'" performs the best on various sets of our data. max\_depth dictates the maximum depth of each decision tree in the Random Forest algorithm. criterion=entropy changes the splitting criterion of the trees to \textit{information gain}.

\begin{table}[h]
    \caption{Accuracy scores of the best performing algorithms with their hyper parameters tuned. These tests are performed on a separate test set, which is entirely different from the earlier validation set.}
    \centering
    \begin{tabular}{|l|p{0.32\linewidth}|p{0.10\linewidth}|p{0.07\linewidth}|}
         \hline
         \textbf{algorithm} & \textbf{parameters\footnotemark[3]} & \textbf{accuracy} & \textbf{AUC} \\
         \hline
         \hline
         Random Forest & max\_depth\_10, criterion="entropy" & \textbf{0.97348} &\textbf{0.996} \\
         Random Forest & min\_samples\_leaf=3, criterion="entropy", max\_depth=20 & 0.87452 & 0.994 \\
         LinearSVC & max\_iter=20000 & 0.92990 & 0.969 \\
         LinearSVC & max\_iter=20000, dual=False, penalty="l1" & 0.92191 & 0.973 \\
         \hline
    \end{tabular}
    
    \label{tab:scores_best}
\end{table}

Thus, instead of our earlier naive approach to detect rotational symmetries as described in Section \ref{wavelet-based}, we can now use our newly trained Random Forest model on each line pair that is found by the algorithm. Any true case will have the centre of its symmetry at the intersection of the reflection symmetry line pair.

\begin{figure}[ht]
    \centering
    \includegraphics[width=0.83\linewidth]{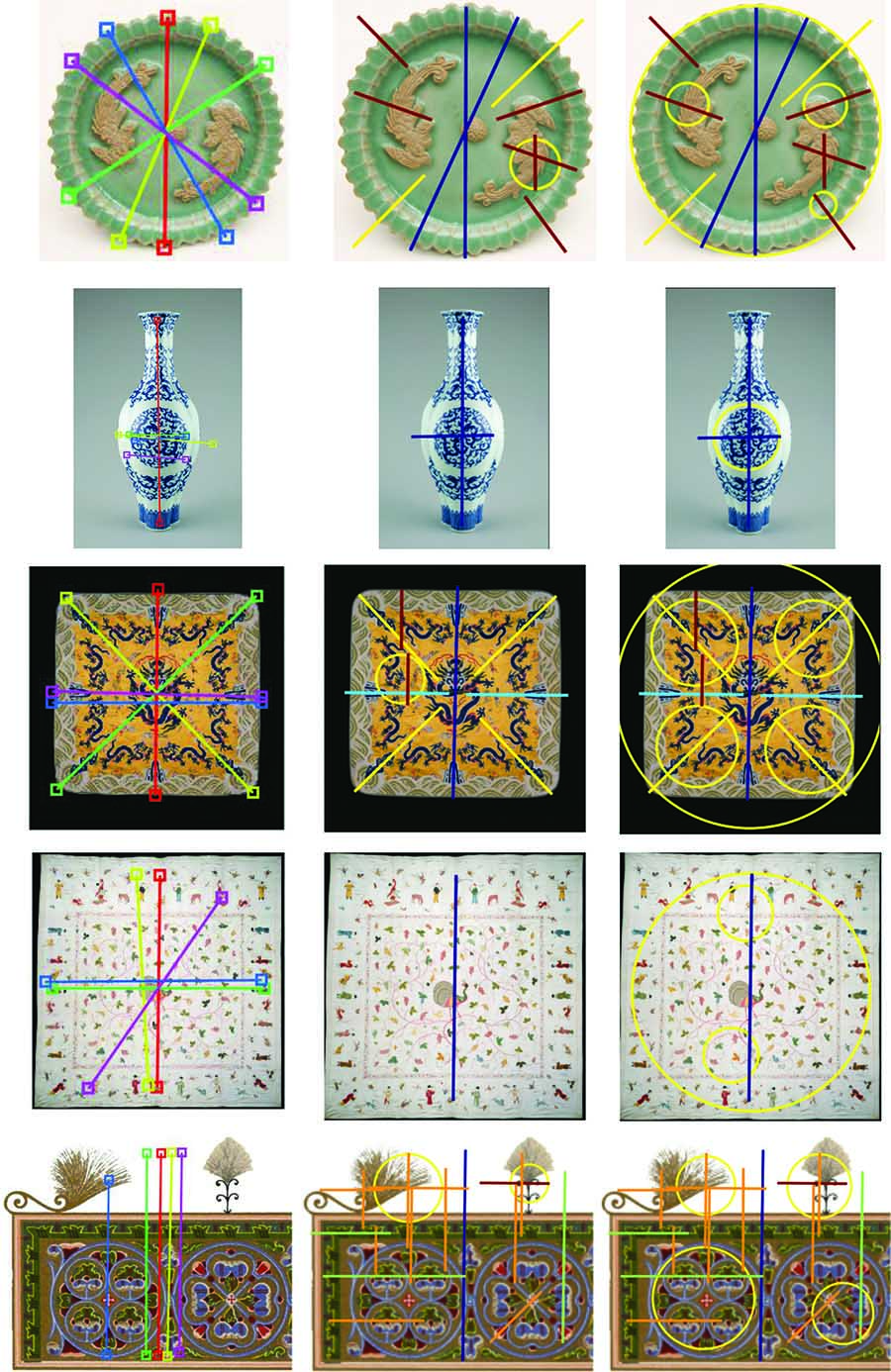}
    \caption{Original WaveletSym on the left, proposed rule-based post processing (Section \ref{wavelet-based}) in the middle, and the proposed machine learning model approach on the right. Rotational symmetries are drawn in yellow. The colour of lines on the right indicates different levels of localised reflection symmetries.
    The proposed machine learning model accurately detects each global rotational symmetry and even some localised symmetries in the 2nd and 5th images. The rule-based approach found no accurate rotational symmetries.}
    \label{fig:All_final}
\end{figure}

\begin{figure}[ht]
    \centering
    \includegraphics[width=0.83\linewidth]{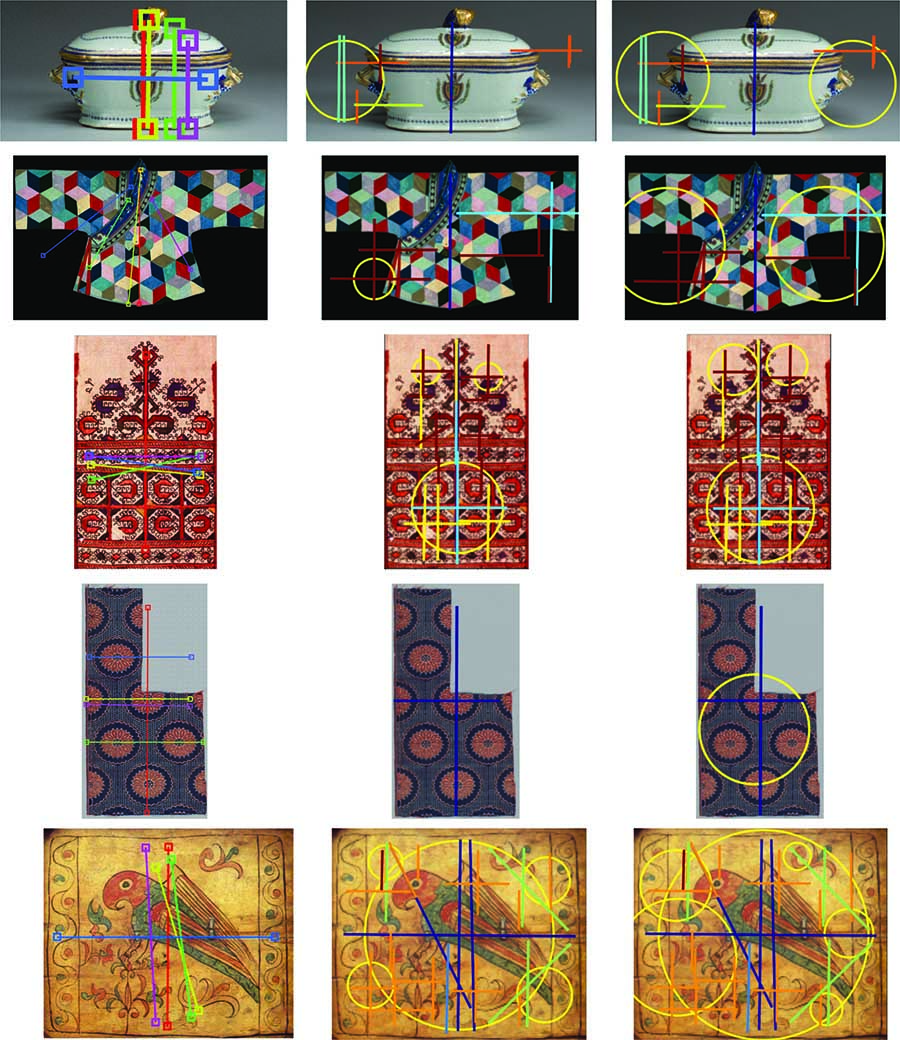}
    \caption{Original WaveletSym on the left, proposed rule-based post processing (Section \ref{wavelet-based}) in the middle, and the proposed machine learning model approach on the right. Rotational symmetries are drawn in yellow. The colour of lines on the right indicates different levels of localised reflection symmetries. Example images where the proposed methods did not perform well.}
    \label{fig:All_final2}
\end{figure}

We found that the proposed machine learning model performs well in identifying rotational symmetries based on two given reflection symmetry lines. However, this does mean that our detection method is still completely reliant on the provided reflection symmetry lines. Due to this, we will now remove fewer symmetry lines as we did in the previous Section (\ref{wavelet-based}) since we can use the model to determine if these lines help detect rotational symmetries or not. After the predictions are made, we will remove the low scoring symmetries as usual. This provides us with a better detection rate for rotational symmetries at the cost of computation time. 

\begin{figure*}[ht]
    \centering
    \subfigure{\includegraphics[height=0.14\linewidth,trim={0mm 0mm 0mm 0mm},clip]{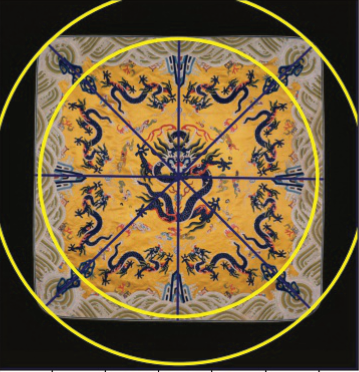}}
    \subfigure{\includegraphics[height=0.14\linewidth,trim={0mm 0mm 0mm 0mm},clip]{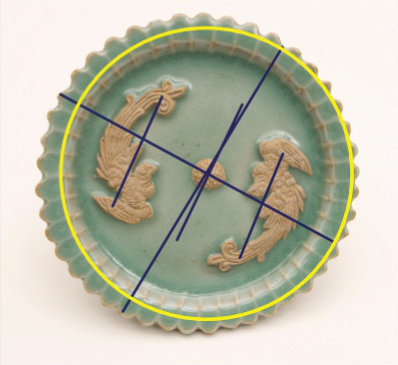}}
    \subfigure{\includegraphics[height=0.14\linewidth,trim={0mm 0mm 0mm 0mm},clip]{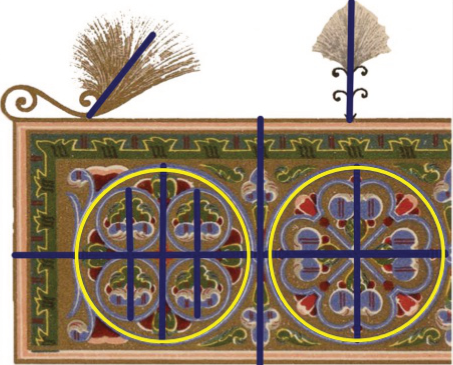}}
    \subfigure{\includegraphics[height=0.14\linewidth,trim={0mm 0mm 0mm 0mm},clip]{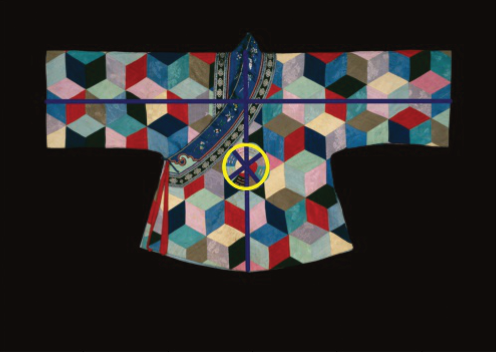}}
    \subfigure{\includegraphics[height=0.14\linewidth,trim={0mm 0mm 0mm 0mm},clip]{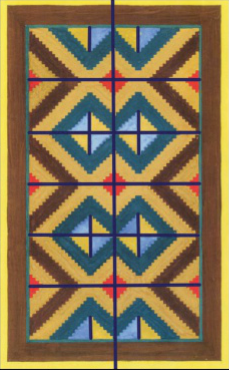}}
    \caption{Blue lines denote human labeled reflection symmetries as ground truths. Yellow circles denote the found rotational symmetries by the proposed machine learning model based on the ground truths.}\label{fig:groundThruthCompare_test}
\end{figure*}

Figures \ref{fig:All_final} and \ref{fig:All_final2} display various images with the original algorithm on the left, the proposed improvements as described in Section \ref{wavelet-based} in the middle, and on the right our improvements with the machine learning model for rotational symmetries.
We can observe that the machine learning model has improved the detection rate of especially the global rotational symmetries. The results do, however, show an increased amount of false positives.
These false positives are mostly caused by wrongly detected reflection symmetries. Figure \ref{fig:groundThruthCompare_test} lists images with their human labeled ground truths (blue lines). When using these ground truths as input we retrieve the resulting rotational symmetries as shown as yellow circles in Figure \ref{fig:groundThruthCompare_test}, showing that the proposed model works well when the symmetry lines are correct.

\begin{figure}[h!]
    \centering
    \includegraphics[width=0.81\linewidth]{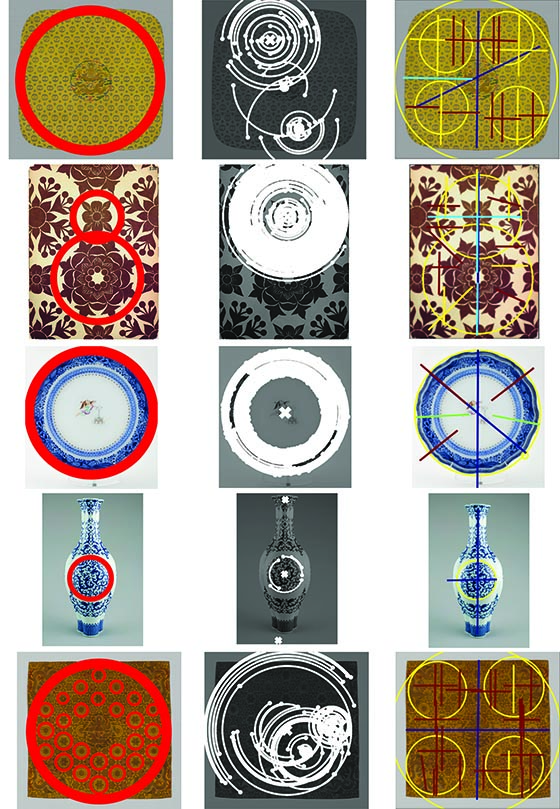}
    \caption{Proposed method (right) compared against Loy and Eklundh \cite{loy2006detecting} (middle). The left column contains the ground truths of the images drawn in red. These images show superior or equal performance of our model over Loy and Eklundh in detecting rotational symmetries.}
    \label{fig:loyCompare1}
\end{figure}

\begin{figure}[h!]
    \centering
    \includegraphics[width=0.81\linewidth]{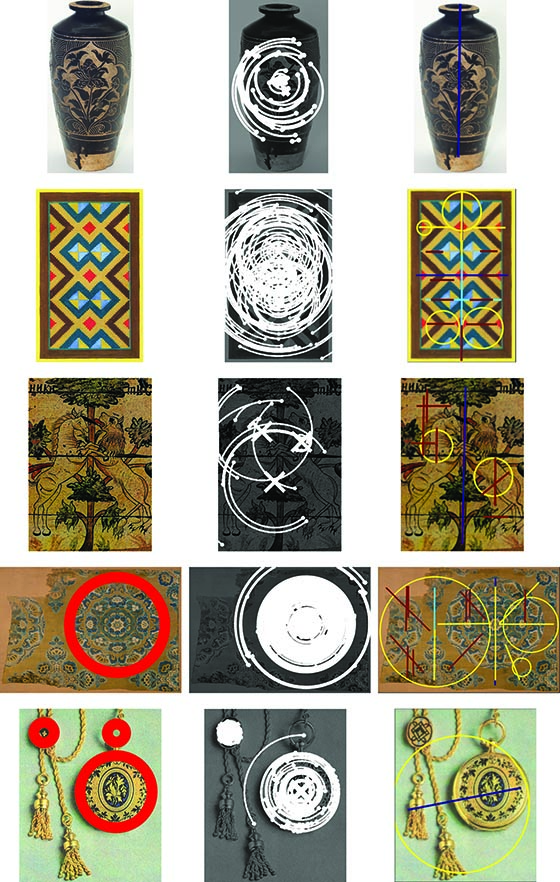}
    \caption{Proposed method (right) compared against Loy and Eklundh \cite{loy2006detecting} (middle). The left column contains the ground truths of the images drawn in red. The first three images contain no rotational symmetries. Our method is better able to detect this. The bottom three images show superior performance of Loy and Eklundh over our model.}
    \label{fig:loyCompare2}
\end{figure}

\paragraph{Comparison with Loy and Eklundh}
In Section \ref{relatedwork} Loy and Eklundh \cite{loy2006detecting} is described as a good benchmark for rotational symmetry detection methods, as it has won two competitions. 
Figure \ref{fig:loyCompare1} and \ref{fig:loyCompare2} show images of our proposed machine learning model against Loy and Eklundh's method.
Our method shows superior performance when it comes to detecting rotational symmetries in images that are completely defined as a rotational symmetry (first, third and fifth image in Figure \ref{fig:loyCompare1}). Rotational symmetries have, by definition, an infinite amount of smaller rotational symmetries with the same centre. Our model will accurately draw the largest encompassing rotational symmetries in images where multiple rotational symmetries exists inside each other. Loy's method does not always find the largest encompassing symmetry or detects a greater symmetry strength in the wrong centre.
Furthermore, in all of our tested images, Loy's method found rotational symmetries in each of them. Our method produces less false positives when there are no rotational symmetries in an image. Examples of this are shown in the first, second and third images in Figure \ref{fig:loyCompare2}. 
The last two images in Figure \ref{fig:loyCompare2} show better performance for Loy and Eklundh. 
Both methods struggle with detecting multiple rotational symmetries. However, Loy's method has a slight edge in this category. Most of the secondary rotational symmetries detected by our method are not accurate, with the current state of reflection symmetry detection.

Since symmetries should often be detected quickly, or in our case, on a large number of images, the speed of each algorithm is an important factor. Table \ref{tab:time_taken} lists the average and median time taken by our approach and Loy and Eklundh's method to detect rotational and reflection symmetries in a set of 10 images.
As is clear from the low median, most rotational symmetries are detected very quickly with our method. However, some outliers do significantly raise the average time taken.
It should be noted that our rotational symmetry detection method requires reflection symmetries as an input. This means the reflection symmetries need to be computed first. On the other hand, Loy and Eklundh's method is able to detect rotational symmetries independently.
Furthermore, during testing, we found that Loy and Eklundh's algorithm required a lot of computational power. Running two instances simultaneously caused our computer to slow down severely during some processing stages.
In contrast, our method was efficiently running on 6 or 7 instances, only being constraint by RAM\footnote{Our computer configuration contained an AMD Ryzen 3600, with 16GB of RAM.}. This means that, depending on the computer configuration, it will be possible to decrease the average detection time by several factors. This greatly benefits cases where lots of images need to be processed, such as with our case for labelling the Ornamika database.

\begin{table}[h]
    \caption{Average and median computation times in seconds of Loy \cite{loy2006detecting} and our approach in both rotation and reflection symmetry detection on a set of 10 images. Our approach requires reflection symmetries to be detected before being able to detect rotational symmetries.} 
    \centering
    \begin{tabular}{|c|c|c|c|c|}
        \hline
        & \multicolumn{2}{c|}{reflection} & \multicolumn{2}{c|}{rotation}\\
        \cline{2-5}
        & our work & Loy \cite{loy2006detecting} & our work & Loy \cite{loy2006detecting} \\
        \hline
        \hline
        average & 171 & 239 & 256 & 95 \\
        median & 109 & 103 & 3 & 38 \\
        \hline
    \end{tabular}
    
    \label{tab:time_taken}
\end{table}

\section{Conclusion}\label{conclusions}
In this paper, we evaluated state-of-the-art symmetry detection algorithm WaveletSym \cite{elawady2017wavelet} on the images of the Ornamika data set \cite{ornamika} and proposed several improvements.
We made alterations to automate the identification of accurately detected symmetries and removing others. 
We proposed a recursive method to apply the algorithm on smaller parts of the image and, in turn, detect more localised symmetries.
We also used the reflection symmetry algorithm to detect rotational symmetries based on a set of rules. 
To improve performance on this rotational symmetry detection method, we proposed a machine learning model based on the Random Forest classifier.
With our model, rotational symmetries are detected with an accuracy of 90\% on the validation set and 97\% on the test set. 
We compared our method of detecting rotational symmetries against Loy and Eklundh's method and found our work to produce more accurate results when a single rotational symmetry is considered.

We can conclude that, with our model, it is possible to detect rotational symmetries with reflection symmetry detection methods. As our model is not bound to a specific algorithm, an improvement in reflection symmetry detection methods will yield improvements to rotational symmetry detection with our model.

\section{Future Perspectives}\label{future}

We found, through observations, lower-resolution images (or smaller parts of images) to produce higher symmetry scores in general with waveletSym. This could imply that a fixed threshold for identifying accurate symmetries is not the right approach. Instead, a higher threshold on each smaller part of images could improve the accuracy.

Because our method relies on reflection symmetry detection methods, its performance should, in theory, increase as reflection symmetry detection methods improve. Because of this, our model can be re-evaluated on any improved reflection detection method that produces fixed $x,y$ coordinates with symmetry scores.

\bibliographystyle{ieeetr}
\bibliography{bibliography}

\begin{thebibliography}{10}

\bibitem{phdthesis}
M.~Elawady, {\em {Reflection Symmetry Detection in Images: Application to
  Photography Analysis}}.
\newblock PhD thesis, Universit{\'e} de Lyon, 2019.

\bibitem{ornamika}
``{Ornamika Database}.'' \url{https://ornamika.com/}.
\newblock Accessed: 2021-06-18.

\bibitem{atallah1984symmetry}
M.~J. Atallah, ``{On Symmetry Detection},'' in {\em Department of Computer
  Science Technical Reports (paper 396), Purdue University}, 1984.

\bibitem{loy2006detecting}
G.~Loy and J.-O. Eklundh, ``{Detecting Symmetry and Symmetric Constellations of
  Features},'' in {\em European Conference on Computer Vision}, pp.~508--521,
  Springer, 2006.

\bibitem{cho2009bilateral}
M.~Cho and K.~M. Lee, ``{Bilateral Symmetry Detection via Symmetry-Growing},''
  in {\em Proceedings of the British Machine Vision Conference}, pp.~4.1--4.11,
  BMVA Press, 2009.
\newblock doi:10.5244/C.23.4.

\bibitem{ming2013symmetry}
Y.~Ming, H.~Li, and X.~He, ``{Symmetry Detection via Contour Grouping},'' in
  {\em 2013 IEEE International Conference on Image Processing}, pp.~4259--4263,
  IEEE, 2013.

\bibitem{sympascal-ke2017srn}
W.~Ke, J.~Chen, J.~Jiao, G.~Zhao, and Q.~Ye, ``{SRN: Side-output Residual
  Network for Object Symmetry Detection in the Wild},'' in {\em Proceedings of
  the IEEE Conference on Computer Vision and Pattern Recognition},
  pp.~1068--1076, 2017.

\bibitem{comp3}
C.~Funk, S.~Lee, M.~R. Oswald, S.~Tsogkas, W.~Shen, A.~Cohen, S.~Dickinson, and
  Y.~Liu, ``2017 iccv challenge: Detecting {S}ymmetry in the {W}ild,'' in {\em
  2017 IEEE International Conference on Computer Vision Workshops (ICCVW)},
  pp.~1692--1701, 2017.

\bibitem{elawady2017wavelet}
M.~Elawady, C.~Ducottet, O.~Alata, C.~Barat, and P.~Colantoni, ``{Wavelet-based
  Reflection Symmetry Detection via Textural and Color Histograms},'' in {\em
  Proceedings, ICCV Workshop on Detecting Symmetry in the Wild}, vol.~3, p.~7,
  2017.

\bibitem{comp1}
I.~Rauschert, K.~Brocklehurst, S.~Kashyap, J.~Liu, and Y.~Liu, ``{First
  Symmetry Detection Competition: Summary and Results},'' {\em The Pennsylvania
  State University, PA, Tech. Rep. CSE11--012}, 2011.

\bibitem{comp2}
J.~Liu, G.~Slota, G.~Zheng, Z.~Wu, M.~Park, S.~Lee, I.~Rauschert, and Y.~Liu,
  ``{Symmetry Detection from Realworld Images Competition 2013: Summary and
  Results},'' in {\em Proceedings of the IEEE Conference on Computer Vision and
  Pattern Recognition Workshops}, pp.~200--205, 2013.

\bibitem{sie2013detecting}
T.~Sie Ho~Lee, S.~Fidler, and S.~Dickinson, ``{Detecting Curved Symmetric Parts
  using a Deformable Disc Model},'' in {\em Proceedings of the IEEE
  international conference on computer vision}, pp.~1753--1760, 2013.

\bibitem{michaelsen2017hierarchical}
E.~Michaelsen and M.~Arens, ``{Hierarchical Grouping using Gestalt
  Assessments},'' in {\em Proceedings of the IEEE International Conference on
  Computer Vision Workshops}, pp.~1702--1709, 2017.

\bibitem{wu2010detecting}
C.~Wu, J.-M. Frahm, and M.~Pollefeys, ``Detecting {Large Repetitive Structures}
  with {Salient Boundaries},'' in {\em European conference on computer vision},
  pp.~142--155, Springer, 2010.

\bibitem{liu2017fusing}
X.~Liu, P.~Lyu, X.~Bai, and M.-M. Cheng, ``{Fusing Image and Segmentation Cues
  for Skeleton Extraction in the Wild},'' in {\em Proceedings of the IEEE
  International Conference on Computer Vision Workshops}, pp.~1744--1748, 2017.

\bibitem{funk2017beyond}
C.~Funk and Y.~Liu, ``{Beyond Planar Symmetry: Modeling Human Perception of
  Reflection and Rotation Symmetries in the Wild},'' in {\em Proceedings of the
  IEEE International Conference on Computer Vision}, pp.~793--803, 2017.

\bibitem{cicconet2014mirror}
M.~Cicconet, D.~Geiger, K.~C. Gunsalus, and M.~Werman, ``Mirror symmetry
  histograms for capturing geometric properties in images,'' in {\em
  Proceedings of the IEEE Conference on Computer Vision and Pattern
  Recognition}, pp.~2981--2986, 2014.

\end{thebibliography}

\end{document}